\pdfoutput=1

\documentclass[11pt]{article}

\usepackage{acl}

\usepackage{times}
\usepackage{latexsym}

\usepackage[T1]{fontenc}

\usepackage[utf8]{inputenc}

\usepackage{microtype}

\usepackage{xcolor}
\definecolor{thedarkblue}{RGB}{0,0,120} %
\definecolor{mydarkblue}{rgb}{0,0.08,0.45} %
\definecolor{darkblue}{rgb}{0,0.08,180}
\colorlet{TufteRed}{red!80!black}
\definecolor{theblue}{RGB}{0,0,180}
\colorlet{thered}{TufteRed}
      
\usepackage{microtype}
\usepackage{balance}

\usepackage{booktabs}
\usepackage{tabularx}

\usepackage{amsmath,amssymb,amsthm}

\newcommand{\eat}[1]{\ignorespaces}
\usepackage{comment}

\newcommand{\journal}[1]{} %

\usepackage{tikz}
\usepackage{verbatim}
\usetikzlibrary{arrows}
\usetikzlibrary{shapes,snakes}
\usetikzlibrary{decorations.pathmorphing} %
\usetikzlibrary{fit}					%
\usetikzlibrary{backgrounds}	%

\usepackage{ragged2e}
\usepackage{multirow}
\usepackage{microtype}
\usepackage{balance}
\usepackage{setspace}

\graphicspath{{./}{./graphics/}}
\newcolumntype{H}{>{\setbox0=\hbox\bgroup}c<{\egroup}@{}}

\newcolumntype{R}[1]{>{\RaggedLeft\arraybackslash}} %
\newcolumntype{L}[1]{>{\RaggedRight\arraybackslash}} %

\newtheorem{Definition}{\hspace{0em}\bfseries{Definition}}

\AtBeginEnvironment{pmatrix}{\setlength{\arraycolsep}{2pt}}

\DeclareMathOperator{\hugeE}{\mbox{\huge\raise-0.3ex\hbox{E}}}
\DeclareMathOperator{\p}{\mathbb{P}}
\DeclareMathOperator{\hugep}{\mbox{\huge\raise-0.3ex\hbox{$\p$}}}

\usepackage{subfigure}
\usepackage{nicefrac}

\DeclareMathAlphabet{\mathbcal}{OMS}{cmsy}{b}{n}
\usepackage{amsmath}
\usepackage{mathrsfs}
\usepackage{comment}

\usepackage{bm}
\usepackage{bbm}
\usepackage{amssymb}
\usepackage{paralist}
\usepackage{graphicx}
\usepackage[normalem]{ulem}
\useunder{\uline}{\ul}{}
\usepackage{booktabs}   %
\usepackage{tabularx}   %
\usepackage{adjustbox}  %
\usepackage[edges]{forest}

\newcommand{\name}{GUI Agent}

\title{GUI Agents: A Survey}

\author{
{\bf Dang Nguyen$^{1}$\thanks{~~Corresponding author: \texttt{dangmn@umd.edu}},
Jian Chen$^2$,
Yu Wang$^3$,
Gang Wu$^4$,
Namyong Park,} \\
{\bf Zhengmian Hu$^4$,
Hanjia Lyu$^5$,
Junda Wu$^6$,
Ryan Aponte$^7$,
Yu Xia$^6$,
Xintong Li$^6$,} \\
{\bf Jing Shi$^4$,
Hongjie Chen$^8$,
Viet Dac Lai$^4$,
Zhouhang Xie$^6$,
Sungchul Kim$^4$,} \\
{\bf Ruiyi Zhang$^4$,
Tong Yu$^4$,
Mehrab Tanjim$^4$,
Nesreen K. Ahmed$^{9}$,
Puneet Mathur$^4$,} \\
{\bf Seunghyun Yoon$^4$,
Lina Yao$^{10}$,
Jihyung Kil$^4$,
Branislav Kveton$^4$,
Thien Huu Nguyen$^3$,} \\
{\bf Trung Bui$^4$,
Tianyi Zhou$^1$,
Ryan A. Rossi$^4$,
Franck Dernoncourt$^4$} \\\\
$^1$University of Maryland,
$^2$State University of New York at Buffalo, \\
$^3$University of Oregon,
$^4$Adobe Research,
$^5$University of Rochester,\\
$^6$University of California, San Diego,
$^7$Carnegie Mellon University,\\
$^8$Dolby Labs,
$^{9}$Cisco Research,
$^{10}$University of New South Wales
}

\begin{document}
\maketitle

\begin{abstract}
Graphical User Interface (GUI) agents, powered by Large Foundation Models, have emerged as a transformative approach to automating human-computer interaction. These agents autonomously interact with digital systems or software applications via GUIs, emulating human actions such as clicking, typing, and navigating visual elements across diverse platforms. Motivated by the growing interest and fundamental importance of GUI agents, we provide a comprehensive survey that categorizes their benchmarks, evaluation metrics, architectures, and training methods. We propose a unified framework that delineates their perception, reasoning, planning, and acting capabilities. Furthermore, we identify important open challenges and discuss key future directions. Finally, this work serves as a basis for practitioners and researchers to gain an intuitive understanding of current progress, techniques, benchmarks, and critical open problems that remain to be addressed.

\end{abstract}

\section{Introduction}
Large Foundation Models (LFMs) have significantly transformed both the landscape of AI research and day-to-day life \citep{bommasani2022opportunitiesrisksfoundationmodels, kapoor2024societalimpactopenfoundation, schneider2024foundation, naveed2024comprehensiveoverviewlargelanguage, wang2024historydevelopmentprincipleslarge}. Recently, we have witnessed a paradigm shift from using LFMs purely as conversational chatbots \citep{touvron2023llama2openfoundation, vicuna2023, dam2024completesurveyllmbasedai} to employing them for performing actions and automating useful tasks \citep{Wang_2024, zhao2023indepthsurveylargelanguage, yao2023reactsynergizingreasoningacting, shinn2023reflexionlanguageagentsverbal, shen2024taskbenchbenchmarkinglargelanguage, cheng2024exploringlargelanguagemodel}. In this direction, one approach stands out: leveraging LFMs to interact with digital systems, such as desktops and mobile phones, or software applications such as a web browser, through Graphical User Interfaces (GUIs) in the same way humans do, for example, by controlling the mouse and keyboard to interact with visual elements displayed on a device's monitor \cite{iong2024openwebagent, hong2023cogagentvisuallanguagemodel, lu2024omniparser, shen2024falcon}.

This approach holds great potential, as GUIs are ubiquitous across almost all computer devices that humans interact with in their work and daily lives. However, deploying LFMs in such environments poses unique challenges, such as dynamic layouts, diverse graphical designs across different platforms, and grounding issues, for instance, fine-grained recognition of elements within a page that are often small, numerous, and scattered \cite{liu2024visualwebbench}. Despite these challenges, many early efforts have shown significant promise \citep{lin2024showuivisionlanguageactionmodelgui, cheng2024seeclick}, and growing interest from major players in the field is becoming evident\footnote{\href{https://www.anthropic.com/news/3-5-models-and-computer-use}{Anthropic}, \href{https://deepmind.google/technologies/project-mariner/}{Google DeepMind}, \href{https://www.bloomberg.com/news/articles/2024-11-13/openai-nears-launch-of-ai-agents-to-automate-tasks-for-users}{OpenAI}}.

Given the immense potential and rapid progress in this field, we propose a unified and systematic framework to categorize the various types of contributions within this space.

\medskip\noindent\textbf{Organization of this Survey.} We begin our survey by clearly defining the term ``GUI Agent,'' 
followed by a formal definition of GUI agent tasks in Section \ref{sec:prelim}.
We then summarize different datasets and environments in Section \ref{sec:benchmarks} to provide readers a clearer picture of the kinds of problem settings currently available. We summarize various GUI agent architectural designs in Section \ref{sec:architecture}, followed by different ways of training them in Section \ref{sec:training}. Lastly, we discuss open problems and future prospects of GUI agent research in Section \ref{sec:open-problems-challenges}.

\section{Preliminaries} \label{sec:prelim}

This section formally defines the term ``GUI agent'' and presents a formalization of GUI agent tasks.

\begin{Definition}[\sc GUI Agent]\label{def:computer_use}
An intelligent autonomous agent that interacts with digital platforms, such as desktops, or mobile phones, through their Graphical User Interface. It identifies and observes interactable visual elements displayed on the device's screen and engages with them by clicking, typing, or tapping, mimicking the interaction patterns of a human user.
\end{Definition}

\paragraph{Problem Formulation.}
GUI agent tasks involve an agent interacting with an environment in a sequential manner. The environment can generally be modeled as a Partially Observable Markov Decision Process (POMDP) \citep{sondik71thesis,hauskrecht00valuefunction}, defined by a tuple $(\mathcal{U}, \mathcal{A}, \mathcal{S}, \mathcal{O}, T)$, where $\mathcal{U}$ is the task space, $\mathcal{A}$ is the action space, $\mathcal{S}$ is the state space (not fully observable to the agent), $\mathcal{O}$ is the observation space, and $T: \mathcal{S} \times \mathcal{A} \to \mathcal{P}(\mathcal{S})$ is a state transition function that maps a state-action pair to a probability distribution over subsequent states.
A GUI agent is a policy $\pi: \Delta^\mathcal{S} \to \mathcal{A}$, where $\Delta^\mathcal{S}$ denotes the probability simplex over the state states. Most commonly, this is implemented using the entire history of past actions and observations. Given a task $u \in \mathcal{U}$, the agent proceeds through a sequence of actions to complete the task. At each step $t$, based on the history of past actions and observations, the policy $\pi$ selects the next action $a \in \mathcal{A}$. The environment then transitions to a new state $s' \in \mathcal{S}$ according to $T$. Depending on the environment's design, the agent may receive a reward $r = R(s, a, s')$, where $R: \mathcal{S} \times \mathcal{A} \times \mathcal{S} \to \mathbb{R}$ is a reward function.

\section{Benchmarks} \label{sec:benchmarks}
GUI agents are developed and evaluated on various platforms, including desktops, mobile phones, and web browser environments. This section summarizes benchmarks for all of these platform types.

When evaluating GUI agents, it is crucial to distinguish between an \textbf{environment} and a \textbf{dataset}. A \textbf{dataset} is a static collection of data point, where each consists of several input features (e.g., a question, a screenshot of the environment, or the current state of the environment) and some output features (e.g., correct answers or actions to be taken). A dataset remains unchanged throughout the evaluation process. 
In contrast, an {\bf environment} is an interactive simulation that represents a real-world scenario of interest. A GUI environment includes the GUI interface of a mobile phone or a desktop.
Unlike datasets, environments are dynamic, actions taken within the environment can alter its state, hence, allowing modeling the problem as Markov Decision Processes (MDPs) or Partially Observable MDPs (POMDPs), with defined action, state, and observation spaces, and a state transition function.

Another critical dimension of the existing benchmarks for GUI agents is the distinction between the open-world and closed-world assumptions. Closed-world datasets or environments presume that all necessary knowledge for solving a task is contained within the benchmark itself. In contrast, open-world benchmarks relax this constraint, allowing relevant information required to complete a task to exist outside the benchmark.

We present a summary of existing GUI agent benchmarks in Table \ref{tab:benchmarks-summary}.

\subsection{Static Datasets}
\subsubsection{Closed-World Datasets.}
RUSS dataset introduces real-world instructions mapped to a domain-specific language (DSL) 
that enables agents to execute web-based tasks with high precision \citep{xu2021grounding}. Similarly, Mind2Web expands the task set to 2000 diverse tasks \citep{deng2024mind2web}, and MT-Mind2Web 
 adapts into conversational settings with multi-turn interactions \citep{deng2024multiturninstructionfollowingconversational}.
In contrast, TURKINGBENCH focuses on common micro tasks in crowdsourcing platforms, featuring a rich mix of textual instructions, multi-modal elements, and complex layouts \citep{xu2024tur}. 
Focusing on visual and textual interplay, VisualWebBench includes OCR, element grounding, and action prediction tasks, which require fine-grained multimodal understanding \citep{liu2024visualwebbench}. 
Similarly, ScreenSpot focuses on GUI grounding for clicking and typing directly from screenshots \citep{cheng2024seeclickharnessingguigrounding}. 
Complementing this, WONDERBREAD extends evaluation to business process management tasks, emphasizing workflow documentation and improvement rather than automation alone \citep{wornow2024multimodal}. 
EnvDistraction dataset explores agent susceptibility to distractions in GUI environments, offering insights into faithfulness and resilience under cluttered and misleading contexts \citep{ma2024cautionenvironmentmultimodalagents}. 
NaviQAte introduces functionality-guided web application navigation, where tasks are framed as QA problems, pushing agents to extract actionable elements from multi-modal inputs \citep{shahbandeh2024naviqate}.

Evaluating on static closed-world datasets is particularly convenient, thanks to their lightweight and ease in setting up compared to environments. They are also especially valuable for fine-grained evaluation, reproducibility, and comparing models under identical conditions. However, they lack the dynamism of real-world applications, as models are tested on fixed data rather than adapting to new inputs or changing scenarios.

\subsubsection{Open-World Datasets.}
While most existing datasets are designed under the closed-world assumption, several datasets do not follow this paradigm. GAIA dataset tests agent integration diverse modalities and tools to answer real-world questions, often requiring web browsing or interaction with external APIs \citep{mialon2023gaiabenchmarkgeneralai}. WebLINX emphasizes multi-turn dialogue for interactive web navigation on real-world sites, enhancing agents' adaptability and conversational skills \citep{lu2024weblinx}.

Evaluation on static open-world datasets balances the ease of evaluation with realism since the agents interact with real-world websites. However, due to the nature of real-world websites, they are often unpredictable and prone to changes, which makes it more challenging to reproduce and compare with prior methods.

\subsection{Interactive Environments} 
\subsubsection{Closed-World Environments.}
Closed-world interactive environments provide controlled and reproducible settings for evaluating agent capabilities.
MiniWoB offers synthetic web tasks requiring interactions with webpages using mouse and keyboard inputs \citep{shi2017world}. 
It focuses on fundamental skills like button clicking and form filling, providing a baseline for evaluating low-level interaction. 
CompWoB extends MiniWoB with compositional tasks, requiring agents to handle multi-step workflows and generalize across task sequences \citep{furuta2023language}.
This introduces dynamic dependencies that reflect real-world complexity. WebShop simulates e-shopping tasks that challenge agents to navigate websites, process instructions, and make strategic decisions \citep{yao2022webshop}.
WebArena advances realism with self-hosted environments across domains like e-commerce and collaborative tools, requiring agents to manage long-horizon tasks \citep{zhou2024webarena}.
VisualWebArena adds multimodal challenges, integrating visual and textual inputs for tasks like navigation and object recognition \citep{koh2024visualwebarena}.
Shifting to enterprise settings, WorkArena evaluates agent performance in complex UI environments, focusing on knowledge work tasks in ServiceNow platform \citep{drouin2024workarenacapablewebagents}. WorkArena++ extends this benchmark by introducing more challenging tasks \citep{DBLP:conf/nips/BoisvertTGCCCCL24}.
ST-WebAgentBench incorporates safety and trustworthiness metrics, assessing policy adherence and minimizing risky actions, critical for business deployment \citep{levy2024stwebagentbenchbenchmarkevaluatingsafety}.
VideoWebArena introduces long-context video-based tasks, requiring agents to understand instructional videos and integrate them with textual and visual data to complete tasks.
It emphasizes memory retention and multimodal reasoning \citep{jang2024videowebarenaevaluatinglongcontext}. In simulated desktop environments, OSWorld \citep{xie2024osworld} provides the first realistic operating system setting for evaluating multimodal GUI agents. Spider2-V \citep{DBLP:conf/nips/CaoLWCFGXZHMXXZ24} builds on this direction by targeting professional-level data science and engineering tasks. BrowserGym \citep{DBLP:journals/corr/abs-2412-05467} develops a unified ecosystem consisting of seven web agent benchmarks for developing and evaluating web agents.

Closed-world environments serve as evaluation platforms that mimic the dynamism of real-world environments while offering stability and reproducibility. However, setting up such benchmarks is often challenging, as they typically require considerable storage space and engineering skills.

\subsubsection{Open-World Environments.}
Open-world interactive environments challenge agents to navigate dynamic, real-world websites with evolving content and interfaces. 
WebVLN introduces a novel benchmark for vision-and-language navigation on websites, requiring agents to interpret visual and textual instructions to complete tasks such as answering user queries \citep{chen2023webvlnvisionandlanguagenavigationwebsites}. 
It emphasizes multimodal reasoning by integrating HTML structure with rendered webpages, setting a foundation for realistic web navigation. 
WebVoyager leverages LLM to perform end-to-end navigation on 15 real websites with diverse tasks \citep{he2024webvoyagerbuildingendtoendweb}. Its multimodal approach integrates screenshots and HTML content, enabling robust decision-making in dynamic online settings. 
AutoWebGLM %
optimizes web navigation through HTML simplification and reinforcement learning \citep{lai2024autowebglmlargelanguagemodelbased}. 
This framework tackles the challenges of diverse action spaces and complex web structures, demonstrating significant improvement in real-world tasks with its AutoWebBench benchmark. MMInA evaluates agents on multihop, multimodal tasks across evolving real-world websites \citep{zhang2024mminabenchmarkingmultihopmultimodal}. The benchmark includes 1,050 tasks requiring sequential reasoning and multimodal integration to complete compositional objectives, such as comparing products across platforms. 
WebCanvas pioneers a dynamic evaluation framework to assess agents in live web environments \citep{pan2024webcanvasbenchmarkingwebagents}. 
Its Mind2Web-Live dataset captures the adaptability of agents to interface changes and includes metrics like key-node-based intermediate evaluation, fostering progress in online web agent research.

Open-world environments are ideal for achieving both realism and dynamism. However, 
getting consistent evaluation and reproducibility is difficult as they evaluate agents on live websites that are subject to frequent changes.

\subsection{Evaluation Metrics} \label{sec:eval}
\subsubsection{Task Completion Metrics.}
The majority of benchmarks use task completion rate as the primary metric to measure GUI agents' performance. However, different papers define task completion differently. Success can be defined as whether an agent successfully stops at a goal state \citep{chen2023webvlnvisionandlanguagenavigationwebsites, zhou2024webarena}, with \citet{zhou2024webarena} programmatically checking if the intended outcome has been achieved (e.g., a comment has been posted, or a form has been completed), or whether the returned results exactly match the ground truth labels \citep{shi2017world, yao2022webshop, koh2024visualwebarena, drouin2024workarenacapablewebagents, levy2024stwebagentbenchbenchmarkevaluatingsafety, mialon2023gaiabenchmarkgeneralai}. 
Another approach is to measure success based on whether an agent completes all required subtasks \citep{lai2024autowebglmlargelanguagemodelbased, zhang2024mminabenchmarkingmultihopmultimodal, pan2024webcanvasbenchmarkingwebagents, furuta2023language, jang2024videowebarenaevaluatinglongcontext, cheng2024seeclickharnessingguigrounding}. 
This approach can be further extended to measure partial success, as shown in \citet{zhang2024mminabenchmarkingmultihopmultimodal}. 
WebVoyager uses GPT-4V to automatically determine success based on the agent's trajectory, reporting a high agreement rate of 85.3\% with human judgments \citep{he2024webvoyagerbuildingendtoendweb}. 
Instead of using a single final-state success metric, WebLINX measures an overall success rate based on aggregated turn-level success metrics across tasks \citep{lu2024weblinx}. 
These turn-level metrics, including Intersection over Union and F1, are computed based on the type of action taken.
Lastly, there are task-specific metrics to measure success, e.g., using ROUGE-L, F1 for open-ended generation \citep{liu2024visualwebbench, xu2024tur, wornow2024multimodal}, accuracy for multiple choice question tasks \citep{liu2024visualwebbench}, Precision and Recall for Standard Operating Procedure validation \citep{wornow2024multimodal}.

\subsubsection{Intermediate Step Metrics.}
While the task completion rate is a single straightforward metric that simplifies evaluation, it fails to provide clear insights into their specific behaviors. Although some fine-grained metrics measure step-wise performance, their scope remains limited.
WebCanvas \citep{pan2024webcanvasbenchmarkingwebagents} evaluates step scores using three distinct targets: URL Matching, which verifies whether the agent navigated to the correct webpage; Element Path Matching, which checks if the agent interacted with the appropriate UI element, such as a button or text box; and Element Value Matching, which ensures the agent inputted or extracted the correct values, such as filling a form or reading text. WebLINX \citep{lu2024weblinx} uses an intent match metric to assess whether the predicted action’s intent aligns with the reference intent. Similarly, Mind2Web \citep{deng2024mind2web} and MT-Mind2Web \citep{deng2024multiturninstructionfollowingconversational} evaluate Element Accuracy by measuring the rate at which the agent selects the correct elements. These systems also measure the precision, recall, and F1 score for token-level operations, such as clicking or typing, and calculate the Step Success Rate, which reflects the proportion of individual task steps completed correctly. While step-wise evaluations provide more fine-grained insight into the agent's performance, it is often challenging to collect reference labels at the step level while also providing enough flexibility to consider different paths to achieve the original tasks.

\subsubsection{Efficiency, Generalization, Safety and Robustness Metrics.} 
Lastly, we summarize additional metrics that evaluate various aspects of GUI agents beyond their raw performance. Existing benchmarks include metrics for efficiency \citep{shahbandeh2024naviqate, chen2023webvlnvisionandlanguagenavigationwebsites, shahbandeh2024naviqate}, generalization across diverse or compositional task settings \citep{furuta2023language}, adherence to safety policies \citep{levy2024stwebagentbenchbenchmarkevaluatingsafety}, and robustness to environmental distractions \citep{ma2024cautionenvironmentmultimodalagents}.

\section{GUI Agent Architectures} 
\label{sec:architecture}
This section focuses on various architectural designs of a GUI agent, which we categorize into four main types: (1) \textbf{Perception}: designs that enable the GUI agent to perceive and interpret observations from its environment; (2) \textbf{Reasoning}: designs related to the cognitive processes of a GUI agent, such as using an external knowledge base for long-term memory access or a world model of the environment to support other modules like planning; (3) \textbf{Planning}: designs related to decomposing a task into subtasks and creating a plan for their execution; and (4) \textbf{Acting}: mechanisms that allow the GUI agent to interact with the environment, including representing actions in natural language using specific templates, JSON, or programming languages as action representations. We present a taxonomy of GUI agent architectures in Figure \ref{fig:agent_architectures}.

\subsection{Perception}

Unlike API-based agents that process structured, program-readable data~\cite{xu2025skill}, GUI agents must perceive and understand the on-screen environment that is designed for human consumption. 
This requires carefully chosen interfaces that allow agents to discover the location, identity, and properties of the interactive elements. 
Broadly, these perception interfaces can be categorized into four types: accessibility-based, HTML/DOM-based, screen-visual-based, and hybrid ones, with each offering different capabilities and posing distinct privacy and implementation considerations.

\subsubsection{Accessibility-Based Interfaces}
Modern mobile and desktop operating systems usually provide accessibility APIs\footnote{\url{https://en.wikipedia.org/wiki/Computer_accessibility}} that expose a semantic hierarchy of UI components, including their roles, labels, and states\footnote{\url{https://developer.apple.com/library/archive/documentation/Accessibility/Conceptual/AccessibilityMacOSX/OSXAXmodel.html}}\footnote{\url{https://developer.apple.com/design/human-interface-guidelines/accessibility}}\footnote{\url{https://learn.microsoft.com/en-us/windows/apps/design/accessibility/accessibility}}. 
GUI agents can utilize accessibility APIs to identify actionable elements and derive semantic cues without relying solely on pixel-based detection. 
These interfaces are resilient to minor layout changes or styling updates; however, their effectiveness depends on proper implementation by developers. 
Accessibility APIs may also be limited when dealing with highly dynamic elements (e.g., custom drawing canvases or gaming environments) and may not natively expose visual content. 
Although these APIs help reduce the complexity of visually parsing the screen, the agent may need additional perception methods for full functionality. 
On the positive side, accessibility-based interfaces typically require minimal sensitive user data, thereby reducing privacy concerns.

\subsubsection{HTML/DOM-Based Interfaces}
For web GUIs, agents frequently utilize the Document Object Model (DOM) to interpret the structural layout of a page. The DOM provides a hierarchical representation of elements, allowing agents to locate targets like buttons or input fields based on tags, attributes, or text content. However, raw HTML data or DOM tree usually has redundant and noisy structure. Various methods are proposed to handle this. Mind2Web \citep{deng2024mind2web} utilizes a fine-tuned small LM to rank the elements in a page before the final prediction of action with a large LM, and WebAgent \citep{gur2024realworldwebagentplanninglong} uses a specialized model HTML-T5 to generate task-specific HTML snippets. 
AutoWebGLM \citep{lai2024autowebglmlargelanguagemodelbased} designs an algorithm to simplify HTML content. While HTML/DOM-based interfaces provide rich structural data, they require careful preprocessing and, in some cases, additional heuristics or trained models to locate and interpret key UI components accurately.

\subsubsection{Screen-visual-based Interfaces}
With advances in computer vision and multimodal LLMs, agents can utilize screen-visual information, such as screenshots, to perceive the on-screen environment.
OmniParser \citep{lu2024omniparser} utilizes an existing multimodal LLM (e.g., GPT-4V) to parse a screenshot into a structured representation of the UI elements.
TAG \citep{DBLP:conf/aaai/XuCWL25} leverages the inherent attention patterns in pretrained MLLMs to improve GUI grounding without the need
for additional fine-tuning.
Cradle \citep{tan2024cradleempoweringfoundationagents}, instead of relying on a single screenshot, processes a video recording (i.e., a sequence of screenshots) to enable more general-purpose computer control.
However, screen-visual-based perception introduces privacy concerns since entire screenshots may contain sensitive information. 
Additionally, computational overhead increases as models must handle high-dimensional image inputs. 
Despite these challenges, such interfaces are crucial for agents operating in environments where high-quality accessibility interfaces and DOM information are unavailable, or environments where dynamic or visual information is crucial, like image or video editing software. A key advantage of this approach is that it requires no application instrumentation, enabling direct deployment across a wide range of applications.

\subsubsection{Hybrid Interfaces}
To achieve robust and flexible performance across diverse environments, many GUI agents employ a hybrid approach \citep{gou2024navigatingdigitalworldhumans, wu2024atlas, DBLP:conf/cvpr/KilSZ00C24}. These systems combine accessibility APIs, DOM data, and screen-visual information to form a more comprehensive understanding of the interface. Leading methods in GUI agent tasks, such as OS-Atlas \citep{wu2024atlas} and UGround \citep{gou2024navigatingdigitalworldhumans}, demonstrate that hybrid interfaces that combine visual and textual inputs can enhance performance. Such approaches also facilitate error recovery, when accessibility or DOM data are incomplete or misleading, the agent can fall back on screen parsing, and vice versa.

\subsection{Reasoning}
WebPilot employs a dual optimization strategy for reasoning \citep{zhang2024webpilot}.
WebOccam improves reasoning by refining the observation and action space of LLM agents \citep{yang2024agentoccamsimplestrongbaseline}.
OSCAR introduces a general-purpose agent to generate Python code from human instructions \citep{wang2024oscar}.
LAST leverages LLMs for reasoning, acting, and planning \citep{zhou2023language}.

\subsection{Planning}
Planning involves decomposing a global task into multiple subtasks that progressively approach the goal state starting from an initial state~\cite{huang2024understanding}. Traditional planning methods, such as symbolic approaches \citep{DBLP:conf/ecai/KautzS92} and reinforcement learning \citep{sutton98reinforcement}, have significant limitations: symbolic methods require extensive human expertise to define rigid system rules and lack error tolerance~\cite{belta2007symbolic, pallagani2022plansformer}, while reinforcement learning demands impractical volumes of training data, often derived from costly environmental interactions~\cite{acharya2023neurosymbolic}. Recent advancements in LLM-powered agents offer a transformative alternative by positioning LLM-powered agents as the cognitive core for planning agents~\cite{huang2024understanding}. When equipping agents with GUIs as the medium, LLM-powered agents can directly interact with nearly all application domains and resources to enhance planning strategies. Based on what application domains/resources agents use for planning, we divide existing works into planning with internal and external knowledge.

\subsubsection{Planning with Internal Knowledge}
Planning with internal knowledge of GUI agents is to leverage the inherent knowledge to reason and think about the potential plans to fulfill the global task goals~\cite{schraagen2000cognitive}.
WebDreamer~\citep{gu2024llmsecretlyworldmodel} uses LLMs to simulate the outcomes of the actions of each agent and then evaluates the result to determine the optimal plan at each step. 
MobA \citep{zhu2024mobatwolevelagentefficient} devises a two-level architecture to power the mobile phone management, with a high level for understanding user commands, tracking history memories and planning tasks, and a low level to act the planned module. 
Agent S \citep{agashe2024agentsopenagentic} introduces an experience-augmented hierarchical planning to perform complex computer tasks.

\subsubsection{Planning with External Knowledge}

Enabling LLM-powered agents to interact with diverse applications and resources through GUIs allows them to leverage external data sources, thereby enhancing their planning capabilities. 
For example, Search-Agent \citep{koh2024treesearchlanguagemodel} combines LLM inference with A* search to explore and backtrack to alternative paths explicitly, AgentQ \citep{putta2024agentqadvancedreasoning} combines LLM with MCTS. 
Toolchain \citep{zhuangtoolchain} models tool planning as a tree search algorithm and incorporates A* search to adaptively retrieve the most promising tool for subsequent use based on accumulated and anticipated costs. 
SGC~\citep{wu2024can} decomposes the query and performs embedding similarity match between the concatenated subquery with the current retrieved task API and each of the existing APIs, and then selects the top one from the existing neighboring APIs. 
Thought Propagation Retrieval~\citep{yu2023thought} prompts LLMs to propose a set of analogous problems and then applies established prompting techniques, like Chain-of-Thought, to derive solutions. 
The aggregation module subsequently consolidates solutions from these analogous problems, enhancing the problem-solving process for the original input. 
Benchmarks like WebShop, Mind2Web, and WebArena~\cite{zhou2023webarena, deng2024mind2web} enable agents to interact with web environments to plan and execute browsing actions for information-seeking tasks.
WMA~\citep{chae2024webagentsworldmodels} utilizes world models to address the mistakes made by LLMs for long-horizon tasks.

\subsection{Acting}

Acting in GUI agents involves translating the agent’s reasoning and planning outputs into executable steps within the GUI environment. Unlike purely text-based or API-driven agents, GUI agents must articulate their actions at a finer granularity—often down to pixel-level coordinates—while also handling higher-level semantic actions such as typing text, scrolling, or clicking on specific elements. Several approaches have emerged:

Those utilizing textual interfaces may only rely on text-based metadata (HTML, accessibility trees) to identify UI elements. For example, WebAgent \citep{gur2024realworldwebagentplanninglong} and Mind2Web \citep{deng2024mind2web} use DOM or HTML representations to locate interactive elements. Similarly, AppAgent \citep{zhang2023appagentmultimodalagentssmartphone} and MobileAgent \citep{wang2024mobileagentautonomousmultimodalmobile} leverage accessibility APIs to identify GUI components on mobile platforms.

However, as highlighted in UGround \citep{gou2024navigatingdigitalworldhumans}, such metadata can be noisy, incomplete, and computationally expensive to parse at every step. 
To overcome these limitations, recent research emphasizes visual-only grounding—mapping textual referring expressions or instructions directly to pixel-level coordinates on a screenshot. 
UGround trains large action models using only screen-level visual inputs. 
OmniParser \citep{lu2024omniparser} also demonstrates how vision-only approaches can parse GUIs without HTML or accessibility data.
Similarly, OS-Atlas \citep{wu2024atlas} leverages large-scale multi-platform training data to achieve universal GUI grounding that generalizes across web, mobile, and desktop platforms. By unifying data sources and action schemas, OS-Atlas showcases the feasibility of a universal approach to action grounding.

\section{\name{} Training Methods} \label{sec:training}
This section summarizes different strategies to elicit the ability to solve agentic tasks in a \name{} agent. We broadly categorize these strategies into two types: (1) \textbf{Prompt-based Methods} and (2) \textbf{Training-based Methods}. Prompt-based methods do not involve the training of parameters; they elicit the ability to solve agentic tasks by providing detailed instructions within the prompt. Training-based methods, on the other hand, involve optimizing the agent's parameters to maximize an objective, such as pretraining, fine-tuning, or reinforcement learning. We present a taxonomy of GUI agent training methods in Figure \ref{fig:agent_training_methods}.

\subsection{Prompt-based Methods}

Prompt-based methods enable GUI agents to exhibit learning and adaptation during inference through carefully designed prompts and interaction mechanisms, without modifying model parameters. 
This learning and adaptation occur as the agent's state evolves by incorporating context from past actions or stored knowledge.

Agent Q \citep{putta2024agentqadvancedreasoning} and OSCAR \citep{wang2024oscar} incorporate self-reflection and self-critique mechanisms via prompts, enabling agents to iteratively improve decision-making by identifying and rectifying errors. 
Auto-Intent \citep{kim2024auto} focuses on unsupervised intent discovery and utilization, extracting intents from interaction histories and incorporating them into future prompts.
Other techniques include state-space exploration in LASER \citep{ma2024laserllmagentstatespace}, state machine in OSCAR \citep{wang2024oscar}, expert development and multi-agent collaboration in MobileExperts \citep{zhang2024mobileexpertsdynamictoolenabledagent}, and app memory in AutoDroid \citep{wen2024autodroid}.

Despite the potential of prompt-based methods, the limited context size of LLMs and the difficulty of designing effective prompts that elicit the desired behavior remain.

\subsection{Training-based Methods}
\subsubsection{Pre-training}
Earlier models for GUI tasks relied on assembling smaller encoder-decoder architectures to address visual understanding challenges due to its ability to learn unified representations from diverse visual and textual data, enhance transfer learning capabilities, and integrate multiple modalities deeply. For example, PIX2STRUCT~\cite{lee2023pix2structscreenshotparsingpretraining} is pre-trained on a screenshot parsing task, which involves predicting simplified HTML representations from screenshots with visually masked regions. It employs a ViT~\cite{dosovitskiy2020image} as the image encoder, T5~\cite{raffel2020exploring} as the text encoder, and a Transformer-based decoder. 

Training of recent GUI agent models often involve the continual pre-training of existing vision large language models on additional large-scale datasets. This step refines the model’s general knowledge and modifies or assembles new neural network modules into the backbone, providing a stronger foundation before fine-tuning on smaller, curated datasets for GUI tasks. VisionLLM \cite{wang2023visionllmlargelanguagemodel} utilizes public datasets to integrate BERT \cite{devlin2018bert} and Deformable DETR \cite{zhu2020deformable} into large language models, focusing on visual question answering tasks centered on grounding and detection. SeeClick \cite{cheng2024seeclick} is built using continual pre-training on Qwen-VL \cite{bai2023qwen} with datasets incorporating OCR-based layout annotation to predict click actions. UGround \cite{gou2024navigatingdigitalworldhumans} uses continual pre-training on the LLaVA-NEXT \cite{liu2024llava} model without its low-resolution image fusion module on a large dataset and synthetic data to align visual elements with HTML metadata for planning and grounding tasks. 

Pre-training is also used to adapt new designs for improved computational efficiency in GUI-related tasks. CogAgent \cite{hong2023cogagentvisuallanguagemodel} employs a high-resolution cross-module to process small icons and text, enhancing its efficiency for GUI tasks such as DOM element generation and action prediction. ShowUI \cite{lin2024showuivisionlanguageactionmodelgui} builds on Qwen2-VL \cite{wang2024qwen2} with a visual-token selection module to improve the computational efficiency for interleaved high-resolution grounding.

\subsubsection{Fine-tuning}
Fine-tuning has emerged as a key strategy to adapt large vision-language models (VLMs) and large language models (LLMs) to the specialized domain of GUI interaction.
Unlike zero-shot or prompt-only approaches, fine-tuning can enhance both the model’s grounding in GUI elements and its ability to execute instructions reliably.

Recent work highlights reducing hallucinations and improving grounding. 
Falcon-UI~\cite{shen2024falcon} fine-tunes on large-scale instruction-free GUI data and then fine-tunes on Android and Web tasks,
achieving high accuracy with fewer parameters. 
VGA~\cite{ziyang2024vga}, through image-centric fine-tuning, 
reduces hallucinations by tightly coupling visual inputs with GUI elements, thus improving action reliability. 
Similarly, UI-Pro~\cite{li2024uipro} identifies a recipe for fine-tuning of VLMs, reducing model size while maintaining state-of-the-art grounding accuracy.

Other methods leverage fine-tuning to incorporate domain-specific reasoning and functionalities, such as functionality-aware fine-tuning for generating human-like interactions~\cite{liu2024make} and alignment strategies to handle multilingual, variable-resolution GUI inputs ~\cite{nong2024mobileflow}. Some methods emphasize autonomous adaptation, such as learning to execute arbitrary voice commands through trial-and-error exploration \cite{pan2023autotask} and learning for cross-platform GUI grounding without structured text \cite{cheng2024seeclick}. Additionally, fine-tuning can specialize models for context-sensitive actions. 
Techniques proposed by \citet{liu2023fill} enable context-aware text input generation, improving coverage in GUI testing scenarios. 
Taken together, these fine-tuning methods demonstrate how careful parameter adaptation, 
data scaling and multimodal alignment can collectively advance the reliability, interpretability, 
and performance of GUI agents.

\subsubsection{Reinforcement Learning}
Reinforcement learning was used in the early text-based agent WebGPT to improve information retrieval of the GPT-3 based model~\cite{nakano2021webgpt}. ~\citet{liu2018reinforcementlearningwebinterfaces} use human demonstrations to constrain the search space for RL, through using \textit{workflows} as a high-level process for the model to complete without specifying the specific details. An example from ~\citet{liu2018reinforcementlearningwebinterfaces} is for the specific process of forwarding a given email, the \textit{workflow} would involve clicking forward, typing in the address, and clicking send. ~\citet{deng2024mind2web} 
use RL based on human demonstrations as the reward signal. While early agents constrained the input and action spaces to only text, recent work has extended to GUI agents. %

WebRL framework uses RL to generate new tasks based on previously unsuccessful attempts as a mitigation for sparse rewards \cite{qi2024webrltrainingllmweb}. Task success is evaluated by an LLM-based outcome reward model (ORM) and KL-divergence is used to prevent significant shifts in policies during curriculum learning.
AutoGLM applies online, curriculum learning, in particular to address error recovery during real-world use and to correct for stochasticity not present in simulators \cite{liu2024autoglm}. 
DigiRL uses a modified advantage-weighted regression (AWR) algorithm for offline learning \cite{peng2019advantageweightedregressionsimplescalable}, but modifies AWR for more stochastic environments by using a simple value function and curriculum learning. 

\section{Open Problems \& Challenges} \label{sec:open-problems-challenges}

\paragraph{User Intent Understanding.}
GUI Agents still struggle to accurately infer user goals across diverse applications, achieving only 51.1\% accuracy on unseen websites \cite{kim2024auto}. Designing models that generalize effectively across varying tasks is crucial, particularly for handling contextual variations in user interactions~\cite{stefanidi2022real} and predicting user behavior in complex interfaces~\cite{gao2024assistgui}. A prospective future research direction is to leverage robust training techniques to enable agents to adapt to new environments with minimal retraining, ultimately providing more seamless and adaptive user experiences. Other promising directions could include training on diverse user interaction datasets and incorporating context-aware learning techniques that utilize historical user actions to better predict intent.

\paragraph{Security and Privacy.}
GUI agents frequently interact with sensitive data such as passwords, confidential documents, and personal credentials, raising serious privacy and security concerns~\cite{he2024emerged, zhang2024large}. These risks are further amplified when agents rely on cloud-based processing, which involves transmitting sensitive information to remote servers. Unauthorized access or incorrect actions could result in severe consequences~\cite{zhang2024privacyasst}. Future research could focus on developing privacy-preserving protocols, such as homomorphic encryption or differential privacy, to ensure data remains secure during both inference and storage. Additional directions may include exploring local processing alternatives and implementing advanced authentication mechanisms to enhance the reliability and safety of GUI agents across diverse environments.

\paragraph{Inference Latency.}
The need to manage complex interactions across diverse applications often conflicts with the requirement for real-time responsiveness. Optimizing model efficiency without compromising accuracy remains a key challenge, particularly when deploying agents in resource-constrained environments. Key issues include minimizing computational overhead, leveraging hardware acceleration, and balancing trade-offs between speed and resource usage. Addressing these challenges calls for lightweight model architectures and adaptive techniques that enable timely, seamless interactions in dynamic GUI settings. Future research could investigate hardware-aware optimization methods, such as quantization and pruning, or efficient decoding strategies like predictive sampling and multi-token prediction, which can significantly reduce latency while preserving system accuracy.

\paragraph{Personalization.}
is a pivotal aspect in the development of GUI agents, aiming to tailor interactions to individual user preferences and behaviors, thereby enhancing satisfaction and efficiency. Recent work~\cite{berkovitch2024identifying} introduced a method for identifying user goals from UI trajectories, enabling agents to infer intentions and proactively assist users based on their interactions with the interface. Future research could explore more sophisticated models that incorporate user feedback to refine personalization strategies, while ensuring trust and compliance with data protection regulations. Additional directions include implementing explicit feedback mechanisms (e.g., thumbs-up/thumbs-down ratings) and developing robust user profiling techniques that integrate behavioral and contextual data to enable more meaningful and adaptive personalization.

\section{Conclusion} \label{sec:conc}
In this survey, we have thoroughly explored GUI Agents, examining various benchmarks, agent architectures, and training methods. Although considerable strides have been made, problems such as intent understanding, security, latency, and personalization remain critical challenges. We hope that this survey is a valuable resource for researchers, offering structure and practical guidance in this rapidly growing and exciting field, and inspiring more work on GUI Agents. The progress in this area has already benefited mankind, enhancing our daily productivity and transforming the way we interact with computers.

\section*{Limitations}  \label{sec:limitations}
We recognize that some studies have explored interactions between LFM-based agents and digital systems through interfaces other than GUIs, such as Command Line Interfaces (CLI) \citep{nguyen2024dynasaur} or Application Programming Interfaces (API). \citep{song2025browsingapibasedwebagents} However, these approaches are relatively limited in scope compared to GUI-based methods. To maintain a focused scope for our survey, we have chosen not to include them in our discussion.

\bibliography{main,Brano}
\bibliographystyle{acl_natbib}

\appendix

\begin{table*}[t]
\small
\centering
\begin{tabularx}{\textwidth}{@{}l l c c X@{}}
\toprule
\textbf{Benchmark} & \textbf{Domain} & \textbf{Type} & \textbf{World} & \textbf{Highlights} \\ 
\midrule
RUSS \citep{xu2021grounding}          & Web            & Dataset & Closed & Map instructions to a DSL for precise web execution \\
Mind2Web \citep{deng2024mind2web}     & Web   & Dataset & Closed & 2 000 diverse single-turn tasks                                  \\
MT-Mind2Web \citep{deng2024multiturninstructionfollowingconversational} & Web & Dataset & Closed & Conversational, multi-turn variant of Mind2Web                    \\
TURKINGBENCH \citep{xu2024tur}        & Crowdsourcing  & Dataset & Closed & Micro-tasks, complex multimodal layouts                          \\
VisualWebBench \citep{liu2024visualwebbench} & Web & Dataset & Closed & OCR, element grounding, action prediction                        \\
ScreenSpot \citep{cheng2024seeclickharnessingguigrounding} & Screenshots & Dataset & Closed & Click / type grounding direct from images                        \\
WONDERBREAD \citep{wornow2024multimodal} & BPM tasks     & Dataset & Closed & Workflow documentation \& improvement                           \\
EnvDistraction \citep{ma2024cautionenvironmentmultimodalagents} & Synthetic GUI & Dataset & Closed & Measures robustness to clutter/distractions                      \\
NaviQAte \citep{shahbandeh2024naviqate} & Web apps      & Dataset & Closed & QA-framed navigation; functionality-guided                       \\
\midrule
GAIA \citep{mialon2023gaiabenchmarkgeneralai} & General   & Dataset & Open  & Open-word multi-modal QA                            \\
WebLINX \citep{lu2024weblinx}        & Live web       & Dataset & Open  & Multi-turn dialogue navigation                                   \\
\midrule
MiniWoB \citep{shi2017world}         & Synthetic web  & Env. & Closed & Low-level mouse/keyboard skills                                  \\
CompWoB \citep{furuta2023language}   & Synthetic web  & Env. & Closed & Compositional, multi-step workflows                              \\
WebShop \citep{yao2022webshop}       & E-commerce     & Env. & Closed & Shopping with instruction following                              \\
WebArena \citep{zhou2024webarena}    & Self-hosted web& Env. & Closed & Long-horizon, multi-domain tasks                                 \\
VisualWebArena \citep{koh2024visualwebarena}& Self-hosted web& Env. & Closed & Adds pixel-level multimodality                                   \\
WorkArena \citep{drouin2024workarenacapablewebagents} & Web, ServiceNow & Env. & Closed & Enterprise knowledge-work UIs                                    \\
WorkArena++ \citep{DBLP:conf/nips/BoisvertTGCCCCL24} & Web, ServiceNow & Env. & Closed & WorkArena with harder tasks                                    \\
BrowserGym \citep{DBLP:journals/corr/abs-2412-05467} & Web & Env. & Closed & Unified gym environment consists of other web agent benchmarks                                     \\
ST-WebAgentBench \citep{levy2024stwebagentbenchbenchmarkevaluatingsafety} & Self-hosted web & Env. & Closed & Safety trustworthiness metrics                      \\
VideoWebArena \citep{jang2024videowebarenaevaluatinglongcontext} & Video + Web & Env. & Closed & Long-context multimodal reasoning                               \\
OSWorld \citep{xie2024osworld}       & Windows GUI    & Env. & Closed & Desktop OS interactions                                          \\
WindowsAgentArena \citep{bonatti2024windowsagentarenaevaluating} & Windows GUI & Env. & Closed & Benchmarks cross-app Windows tasks                               \\
\midrule
WebVLN \citep{chen2023webvlnvisionandlanguagenavigationwebsites} & Live web & Env. & Open & Vision-language navigation                                        \\
WebVoyager \citep{he2024webvoyagerbuildingendtoendweb} & 15 live sites & Env. & Open & End-to-end nav; HTML + screenshots                                \\
AutoWebBench \citep{lai2024autowebglmlargelanguagemodelbased} & Live web & Env. & Open & RL finetuning, HTML simplification                                \\
MMInA \citep{zhang2024mminabenchmarkingmultihopmultimodal} & Live web & Env. & Open & Multihop, multimodal objectives                                   \\
WebCanvas \citep{pan2024webcanvasbenchmarkingwebagents} & Live web & Env. & Open & Dynamic eval; interface-change resilience                         \\
\bottomrule
\end{tabularx}
\caption{Benchmarks for GUI-agent research discussed in Section~\ref{sec:benchmarks}.  
“Type” distinguishes static \emph{datasets} from interactive \emph{environments}; “World” marks closed- vs. open-world assumptions.}
\label{tab:benchmarks-summary}
\end{table*}

\begin{table*}[t]
\centering
\footnotesize
\resizebox{\textwidth}{!}{%
\begin{tabular}{llll}
\toprule
\textbf{Perception Modality} & \textbf{Data Type}                               & \textbf{Key Advantages}                                                                                                                & \textbf{Key Limitations}                                                                                                                         \\
\midrule
Accessibility-Based          & Structured hierarchy (accessibility APIs)        & \begin{tabular}[c]{@{}l@{}}1) Offers semantic roles/labels\\ 2) Resilient to minor layout changes\\ 3) Lower privacy risk\end{tabular} & \begin{tabular}[c]{@{}l@{}}1) Requires correct developer implementation\\ 2) May not handle highly dynamic or custom-drawn elements\end{tabular} \\
HTML/DOM-Based               & Hierarchical data (DOM tree)                     & \begin{tabular}[c]{@{}l@{}}1) Rich structural information for web-based UIs\\ 2) Directly targets interface elements\end{tabular}      & \begin{tabular}[c]{@{}l@{}}1) HTML can be noisy/redundant\\ 2) Needs careful preprocessing (e.g., snippet extraction, heuristics)\end{tabular}   \\
Screen-Visual-Based          & Pixel data (screenshots)                         & \begin{tabular}[c]{@{}l@{}}1) Universal approach (no reliance on APIs)\\ 2) Handles custom visuals or games\end{tabular}               & \begin{tabular}[c]{@{}l@{}}1) Higher computational overhead\\ 2) Potential privacy concerns (full screenshot capture)\end{tabular}               \\
Hybrid (Multiple Modalities) & Combination (e.g., accessibility + DOM + Screen) & \begin{tabular}[c]{@{}l@{}}1) More robust to missing/incomplete data\\ 2) Better coverage in complex or dynamic tasks\end{tabular}     & \begin{tabular}[c]{@{}l@{}}1) Increased system complexity\\ 2) Requires synchronizing data from multiple modalities\end{tabular}                 \\
\bottomrule
\end{tabular}%
}
\caption{Overview of Perception Modalities}
\label{tab:perception_modalities}
\end{table*}

\begin{table*}[t]
\centering
\small
\resizebox{\textwidth}{!}{%
\begin{tabular}{lll}
\toprule
\textbf{Modality}   & \textbf{Typical Scenarios}                                                                                               & \textbf{Example References}                      \\ \midrule
Accessibility-Based & \begin{tabular}[c]{@{}l@{}}- Desktop/mobile apps with accessibility layers\\ - Automated UI testing/checks\end{tabular}  & OS-based accessibility APIs, Official guidelines \\
HTML/DOM-Based      & \begin{tabular}[c]{@{}l@{}}- Web automation tasks (form-filling, data entry)\\ - Web scraping/search\end{tabular}        & Mind2Web, WebAgent, AutoWebGLM                   \\
Screen-Visual-Based & \begin{tabular}[c]{@{}l@{}}- Image-centric or game UIs\\ - Environments with no structured metadata\end{tabular}         & OmniParser                                       \\
Hybrid              & \begin{tabular}[c]{@{}l@{}}- Complex multi-step tasks\\ - High-value scenarios (e.g., financial dashboards)\end{tabular} & OS-Atlas, UGround                                \\ \bottomrule
\end{tabular}%
}
\caption{Typical Usage Scenarios}
\label{tab:usage_scenarios}
\end{table*}

\begin{figure*}[t]
\centering
\footnotesize
    \begin{forest}
        for tree={
            forked edges,
            draw,
            rounded corners,
            node options={align=center,},
            s sep=6pt,
            calign=center,
            grow=east,
            reversed=true,
            anchor=base west,
            parent anchor=east,
            child anchor=west,
            base=left,
            font=\small,
            minimum width=2.5em,
          },
          where level=1{text width=5em,fill=blue!10}{},
          where level=2{text width=5em,fill=pink!30}{},
        [GUI Agent Architectures, fill=gray!20
            [Perception, text width=50pt, for tree={
                calign=child edge, calign child=(n_children()+1)/2,
            }
                [Accessibility-Based Interfaces, text width=50pt, for tree={
                    calign=child edge, calign child=(n_children()+1)/2,
                }
                    [
                        {
                            OSWorld~\citep{xie2024osworld}
                        }, text width=205pt
                    ]
                ]
                [HTML/DOM-Based Interfaces, text width=50pt, for tree={
                    calign=child edge, calign child=(n_children()+1)/2,
                }
                    [
                        {
                            Mind2Web~\citep{deng2024mind2web}, WebAgent~\citep{gur2024realworldwebagentplanninglong}, AutoWebGLM~\citep{lai2024autowebglmlargelanguagemodelbased}
                        }, text width=205pt
                    ]
                ]
                [Screen-visual-based Interfaces, text width=50pt, for tree={
                    calign=child edge, calign child=(n_children()+1)/2,
                }
                    [
                        {
                            OmniParser~\citep{lu2024omniparser}, TAG~\citep{DBLP:conf/aaai/XuCWL25}, Cradle~\citep{tan2024cradleempoweringfoundationagents}
                        }, text width=205pt
                    ]
                ]
                [Hybrid Interfaces, text width=50pt, for tree={
                    calign=child edge, calign child=(n_children()+1)/2,
                }
                    [
                        {
                            UGround~\citep{gou2024navigatingdigitalworldhumans}, OS-Atlas~\citep{wu2024atlas}
                        }, text width=205pt
                    ]
                ]
            ]
            [Reasoning, text width=50pt, for tree={
                calign=child edge, calign child=(n_children()+1)/2,
            }
                [
                    {
                        WebPilot~\citep{zhang2024webpilot}, WebOccam~\citep{yang2024agentoccamsimplestrongbaseline}, OSCAR~\citep{wang2024oscar}, LAST~\citep{zhou2023language}
                    }, text width=205pt, fill=white
                ]
            ]
            [Planning, text width=50pt, for tree={
                calign=child edge, calign child=(n_children()+1)/2,
            }
                [Planning with Internal Knowledge, text width=50pt, for tree={
                    calign=child edge, calign child=(n_children()+1)/2,
                }
                    [
                        {
                            WebDreamer~\citep{gu2024llmsecretlyworldmodel}, MobA~\citep{zhu2024mobatwolevelagentefficient}, Agent S~\citep{agashe2024agentsopenagentic}
                        }, text width=205pt
                    ]
                ]
                [Planning with External Knowledge, text width=50pt, for tree={
                    calign=child edge, calign child=(n_children()+1)/2,
                }
                    [
                        {
                            Search-Agent~\citep{koh2024treesearchlanguagemodel}, AgentQ~\citep{putta2024agentqadvancedreasoning}, Toolchain~\citep{zhuangtoolchain}, SGC~\citep{wu2024can} Thought Propagation Retrieval~\citep{yu2023thought}, WebShop~\citep{yao2022webshop}, Mind2Web~\citep{deng2024mind2web}, WebArena~\citep{zhou2024webarena}, WMA~\citep{chae2024webagentsworldmodels}
                        }, text width=205pt
                    ]
                ]
            ]
            [Acting, text width=50pt, for tree={
                calign=child edge, calign child=(n_children()+1)/2,
            }
                [
                    {
                        WebAgent~\citep{gur2024realworldwebagentplanninglong}, Mind2Web~\citep{deng2024mind2web}, AppAgent~\citep{zhang2023appagentmultimodalagentssmartphone}, MobileAgent~\citep{wang2024mobileagentautonomousmultimodalmobile}, UGround~\citep{gou2024navigatingdigitalworldhumans} OmniParser~\citep{lu2024omniparser}, OS-Atlas~\citep{wu2024atlas}
                    }, text width=205pt, fill=white
                ]
            ]
        ]
    \end{forest}
    \caption{Taxonomy of GUI agent architectures.}
    \label{fig:agent_architectures}
\end{figure*}

\begin{figure*}[t]
\centering
\footnotesize
    \begin{forest}
        for tree={
            forked edges,
            draw,
            rounded corners,
            node options={align=center,},
            s sep=6pt,
            calign=center,
            grow=east,
            reversed=true,
            anchor=base west,
            parent anchor=east,
            child anchor=west,
            base=left,
            font=\small,
            minimum width=2.5em,
          },
          where level=1{text width=5em,fill=blue!10}{},
          where level=2{text width=5em,fill=pink!30}{},
        [GUI Agent Training Methods, fill=gray!20
            [Prompt-based Methods, text width=50pt, for tree={
                calign=child edge, calign child=(n_children()+1)/2,
            }
                [
                    {
                        Agent Q~\citep{putta2024agentqadvancedreasoning}, OSCAR~\citep{wang2024oscar}, Auto-Intent~\citep{kim2024auto}, AutoDroid~\citep{wen2024autodroid}, LASER~\citep{ma2024laserllmagentstatespace}, MobileExperts~\citep{zhang2024mobileexpertsdynamictoolenabledagent}
                    }, text width=205pt, fill=white
                ]
            ]
            [Training-based Methods, text width=50pt, for tree={
                calign=child edge, calign child=(n_children()+1)/2,
            }
                [Pre-training, text width=50pt, for tree={
                    calign=child edge, calign child=(n_children()+1)/2,
                }
                    [
                        {
                            PIX2STRUCT~\citep{lee2023pix2structscreenshotparsingpretraining}, VisionLLM~\citep{wang2023visionllmlargelanguagemodel},
                            SeeClick~\citep{cheng2024seeclick}, UGround~\citep{gou2024navigatingdigitalworldhumans},
                            CogAgent~\citep{hong2023cogagentvisuallanguagemodel}, ShowUI~\citep{lin2024showuivisionlanguageactionmodelgui}
                        }, text width=205pt
                    ]
                ]
                [Fine-tuning, text width=50pt, for tree={
                    calign=child edge, calign child=(n_children()+1)/2,
                }
                    [
                        {
                            Falcon-UI~\citep{shen2024falcon}, VGA~\citep{ziyang2024vga}, UI-Pro~\citep{li2024uipro},
                            MAKE~\citep{liu2024make}, MobileFlow~\citep{nong2024mobileflow},
                            AutoTask~\citep{pan2023autotask}, SeeClick~\citep{cheng2024seeclick}, FILL~\citep{liu2023fill}
                        }, text width=205pt
                    ]
                ]
                [Reinforcement Learning, text width=50pt, for tree={
                    calign=child edge, calign child=(n_children()+1)/2,
                }
                    [
                        {
                            WebGPT~\citep{nakano2021webgpt}, Workflow RL~\citep{liu2018reinforcementlearningwebinterfaces},
                            Mind2Web~\citep{deng2024mind2web}, WebRL~\citep{qi2024webrltrainingllmweb},
                            AutoGLM~\citep{liu2024autoglm}, DigiRL~\citep{bai2024digirltraininginthewilddevicecontrol}
                        }, text width=205pt
                    ]
                ]
            ]
        ]
    \end{forest}
    \caption{Taxonomy of GUI agent training methods.}
    \label{fig:agent_training_methods}
\end{figure*}

\end{document}